\documentclass{article}

\usepackage{cite}
\usepackage{microtype}
\usepackage{graphicx}
\usepackage{subfigure}
\usepackage{amsmath}
\usepackage{amssymb}
\usepackage{multirow}
\usepackage{booktabs} 

\usepackage[accepted]{icml2020}

\icmltitlerunning{Lifelong Learning Strategy}

\begin{document}

\twocolumn[
\icmltitle{Lifelong Learning Strategy: Self-Memory Supervising and
			\\ Dynamically Growing Process}



\icmlsetsymbol{equal}{*}

\begin{icmlauthorlist}
\icmlauthor{Youcheng Huang}{}
\icmlauthor{Chenwei Tang}{}
\icmlauthor{Jundong Zhou}{}
\icmlauthor{Chunxin Yang}{}

\end{icmlauthorlist}

\icmlkeywords{Lifelong learning, Self-memory supervision, Selective activation, Dynamically growing process}

\vskip 0.3in
]



\begin{abstract}
From childhood to youth, human gradually come to know the world. But for neural networks, this growing process seems difficult. Trapped in catastrophic forgetting, current researchers feed data of all categories to a neural network which remains the same structure in the whole training process. We compare this training process with human learing patterns, and find two major conflicts. In this paper, we study how to solve these conflicts on generative models based on the conditional variational autoencoder(CVAE) model. To solve the uncontinuous conflict, we apply memory playback strategy to maintain the model's recognizing and generating ability on invisible used categories. And we extend the traditional one-way CVAE to a circulatory mode to better accomplish memory playback strategy. To solve the `dead' structure conflict, we rewrite the CVAE formula then are able to make a novel interpretation about the funtions of different parts in CVAE models. Based on the new understanding, we find ways to dynamically extend the network structure when training on new categories. We verify the effectiveness of our methods on MNIST and Fashion MNIST and display some very insteresting results.
\end{abstract}

\section{Introduction}

Since childhood, humans have been constantly learning to recognize the world, to make new memories and to recognize new objects. Humans take this continual learning ability for granted, but it is difficult for current deep learning models to do so. Trapped in catastophic forgetting, most current development classfication and generation methods require a carefully designed model that huge enough to learn all categories' knowledges before the training start, and to feed data with all categories with the model. These kind of models and training process have two major conflicts when comparing to human learning patterns:
\begin{itemize}
\item \textbf{Uncontinuous conflict:}when new objects emerge, human gradually learn new knowledges. However, deep learning models learn knowledges in a label-limited object sets.
\item \textbf{`Dead' structure conflict:}with intelligence developing, neurons in human brains gradually change themselves and become richer. However, deep learning models have `dead' stuctures staying the same.
\end{itemize}
These two conflicts make current deep learning methods less intelligent.

In the continual learning process, the researchers will feed sequential data ordered in categories to train the model, and keep already used categories unavailable as the training process goes on. Catastrophic forgetting is when, without any contraint, the model will no longer hold the knowledges about previous categories\cite{MCCLOSKEY1989109} when it is training on a new category. Note that the image generation tasks are more complex and general than classification tasks. This paper discusses to continually train generation models. More specifically, the continual training of label-conditioned self-encoding models. We pick up conditional variational autoencoder(CVAE)\cite{NIPS2016_5775} because it is simple, not explored yet and has meaningful mathematical formulas.  According to the categories order, we gradually feed each category to train a CVAE network in which applying the self-memory supervision and circulatory mode to solve the first conflict and equipping CVAE with dynamically growing structure to sovle the second conflict. Finally, the network can generate corresponding data through Gaussian noises and condition inputs.

About the open problem of continual learning, recent researchers seek for breakthoughs from two aspects. One is to find better directions of optimizing parameters\cite{Kirkpatrick3521}\cite{DBLP:journals/corr/abs-1901-11356}\cite{DBLP:journals/corr/abs-1710-10628}\cite{DBLP:journals/corr/abs-1903-03511} and the other is based on memory playback\cite{DBLP:journals/corr/abs-1809-02058}\cite{DBLP:journals/corr/SeffBSL17} or knowledge distillation methods\cite{Zhai_2019_ICCV}. The researches focused on parameter optimization have made a deep investigation on the mathematical properties of neural networks. Parameter regularization\cite{DBLP:journals/corr/SeffBSL17} can partly solve the problem of continually training discrimination and generation models, while poor perfermance is observed\cite{DBLP:journals/corr/abs-1809-02058}. And it makes the already unnatural neural networks more elusive because we can't expect our brain neurons to consider such a fine mathematical problem when making changes. Conditional-GAN(CGAN) based on the memory playback and the models based on the usage of knowledge distillation obtain good results. CGANs\cite{DBLP:journals/corr/abs-1809-02058}\cite{DBLP:journals/corr/SeffBSL17} use the conditional batch normalization\cite{DBLP:conf/iclr/DumoulinSK17} to extend its ability to generate label-conditioned images. Although solving the first conflict, the network stucture stays the same and the second conflict remains. Methods using knowledge distillation\cite{Zhai_2019_ICCV} need retrain a new randomly initialized model when new categories come. This retraining process make poor use of old knowledges. The most natural way is to extend the trained model dynamically.

We are inspired by the success of CVAE\cite{NIPS2016_5775} because of its simpleness and highly refined formulas. However, the CVAE formulas ignore the lasting effect of differences across categories when designing its objective function. It assumes that all categories images can be encoded to the latent representation following the same Gaussian distribution, but the significant differences between categories in the actual situation make this assumption unreasonable. We rewrite the CVAE fomulas to supplement the above ignorance, and based on which we can make a novel interpretation about the role of CVAE's condition input and find its selective activation function. Then we come to propose a dynamically growing CVAE structure to solve the second conflict. At the same time, applying memory playback and extending the one-way CVAE model to a circulatory mode allow our methods to solve the first conflict as well. In summary, our contributions are as follows:
\begin{itemize}
\item Rewrite the CVAE formulas, and provide a novel perspective for understanding the function of condition input and other parts of CVAE model.
\item Utilize the CVAE to accomplish memory playback strategy, and extend it to a circulatory mode, providing a new method to solve the uncontinuous conflict.
\item Equip the traditional CVAE model with dynamically growing structure, which provides a novel way to solve the `dead' structure conflict.
\item Conduct experiements on single-pattern MNIST dataset and on multi-patters Fashion-MNIST dataset, whicih verify the effectiveness of our methods and the correctness of our understanding about CVAE.
\end{itemize}

\section{Related Work}

\subsection{Conditional-VAE}

Variational AutoEncodeer(VAE)\cite{Kingma2013AutoEncodingVB} is an important kind of generative model proposed in 2013. In VAE, the visual data $x$ we can observe is assumed related to a latent vector $z$. The mapping function $x \rightarrow z$ is the recognition function $q_\phi (z|x)$, as well as the encoder in VAE. And $z \rightarrow x$ is the generative function $p_\theta (x|z)$, which is the decoder. VAE is designed to fit those two functions. It is now widely used to generate images, and when the generation model is trained, we can generate visual data $\hat{x}$ with different shapes known as datas' manifold\cite{DBLP:journals/corr/RezendeMW14}. Unlike generative adversarial networks(GANs)\cite{DBLP:journals/corr/GoodfellowPMXWOCB14}\cite{DBLP:conf/nips/ChenCDHSSA16}, we know the density function(PDF) of the latent representation(or, we set it), while GANs dose not know such distribution.

VAE is a beyesian model, and the CVAE structure varies as different conditional probabilities are used\cite{NIPS2014_5352}\cite{10.1007/978-3-319-46478-7_51}\cite{Tang2017MultiEntityDL}. Like developing GAN to conditional-GAN\cite{DBLP:journals/corr/MirzaO14}, the condition inforamtion is treated as part of the encoder's inputs or decoder's latent inputs in different CVAEs. So kinds of images can be generated according to given condition\cite{DBLP:conf/cvpr/MishraRMM18}\cite{DBLP:conf/cvmp/DortaVACPS17}. Pandey et al.\cite{DBLP:journals/corr/PandeyD16} discussed how different categories will influence the encoding and decoding of the latent representaion. They thought the condition information (i.e., different categories) is independent with the latent representation. In contrast, they also proposed conditional multimodal autoencoder(CMMA) trying to generate latent vector directly from condition information.

\subsection{Continual Learning on Generative Tasks}

About generative tasks, relatively less work studies the problem of continual learning and catastrophic fogetting. Seff et al. first introduced continual generative modeling and the idea of memory playback\cite{DBLP:journals/corr/SeffBSL17}. Their approach incorporated elastic weight consolidation(EWC) into the loss function of GANs. Wu et al. explored the idea of memory playback futher and well accomplished the label-conditioned images generating tasks\cite{DBLP:journals/corr/abs-1809-02058}. They used conditional batch normalization on CGAN to extend their model's ability to generate new categories images while remaining its structures. But their model presented limited capability in remembering previous categories.

Knowledge distillation is employed by Mengyao Zhai et al. to accomplish image-conditioned and label-conditioned generating tasks\cite{Zhai_2019_ICCV}. They use the cVAE-GAN model and make the networks learn new condition information by adding one more prior condition input (e.g., another size-fixed image in the image-conditioned model). They also used cLR-GAN model which applying condition information in the similar way. Both their model don't need to change their structures. And when training on a new category,  a new random initialized model is needed. By forcing the old model and new model to generate the same auxiliary data, they can partly solve the catastrophic fogetting.

In this paper, we also apply the idea of memory playback. We introduce a new continual learning method which only contains a simple CVAE model. When a new category comes, our model can extend itself while remaining the trained parameters, and growing new `neurons' (i.e., parameters) to learn new knowledges. We further illustrate its mathematical properties and find those categories-related parameters are updating independently, thus our model can distinct and remember different knowledges well. By applying our method, both two conflicts are rationally solved.

\section{Approach}

In this section, we will describe the proposed methods in details. We make discussions about the selective activation function of the One-Hot condition vector in CVAE based on the rewritten formulas.
The understanding of the encoder and decoder will also be discussed. After the discussions, we naturally propose a dynamically growing structure. Finally, we discuss how to introduce the circulatory mode to better explore self-memory supervision.
	
\subsection{CVAE Formulas and Work Principles}

In original VAE proposed by Diederik P.Kingma and Max Welling, the visual data $x$ is transfered from a latent reprensentation $z$. They build models to fit the transformation function by maximizing following variation bound:
\begin{equation}\label{key}
	-D_{KL}(q_\phi(z|x)||p_\theta(z))+\mathbb{E}_{q_\phi(z|x)}[logp_\theta(x|z)]
\end{equation}
And equation 2 is the corresponding original CVAE's variation bound. 
\begin{equation}\label{key}
-D_{KL}(q_\phi(z|x, y)||p_\theta(z))+\mathbb{E}_{q_\phi(z|x)}[logp_\theta(x|z, y)]
\end{equation}
The KL divergence in equation 2 can be minimized under the assumption that $z\sim \mathcal{N}(0, 1)$. This assumption assumes that all categories can be encoded into a latent representation following the same distribution, while ignoring the lasting effect of categorical differences. We think that this ignoracne causes fatal misunderstanding about CVAE structures. So we rewrite the condition variation bound as following\footnote{Details are provided in supplementary material.}:
\begin{equation}\label{key}
\begin{aligned}
&-D_{KL}(q_{\phi_z}(z|x)||p_\theta (z|x))
-D_{KL}(q_{\phi_y}(y|x)||p_\theta (y|x))\\
&+\mathbb{E}_z[logp_\theta (x|z, y)]
\end{aligned}
\end{equation}

In the rewritten formula, we separate the latent representation into two independent parts: `mixed' part $q_{\phi_z}(z|x)$ and conditional part $q_{\phi_y}(y|x)$, then two KL divergences are observed. Applying the traditional CVAE's assumption on $z$ to $q_{\phi_z}(z|x)$, we can minimize the first divergence. However, the second divergence is unoptimizable because we can't know or make resonable assumptions about $q_{\phi_y}(y|x)$.

But we are shocked with the success of CVAE which has above ignorance, so we need to explain how it works.
\begin{figure}[!h]
	\vskip 0.2in
	\begin{center}
		\centerline{\includegraphics[width=0.667\columnwidth]{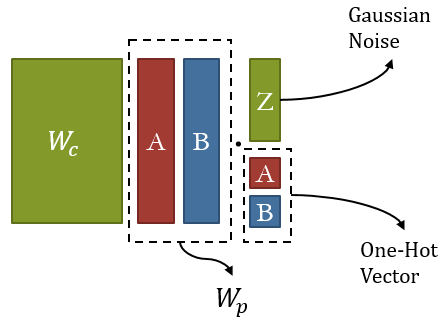}}
		\caption{$1^{st}$ MLP decoder layer with a Gassian noise and a One-Hot vector input.}
		\label{key}
	\end{center}
	\vskip -0.2in
\end{figure}

\begin{figure*}[htb]
	\vskip 0.2in
	\begin{center}
		\centerline{\includegraphics[width=\columnwidth]{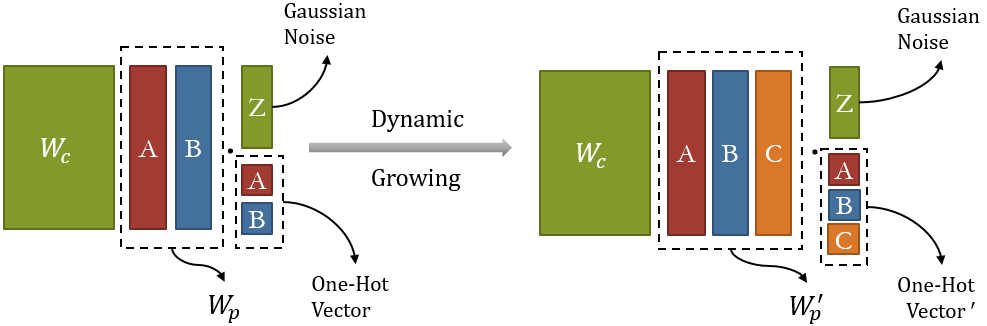}}
		\caption{Dynamically growing process of the $1^{st}$ MLP decoder layer.}
		\label{key}
	\end{center}
	\vskip -0.2in
\end{figure*}
In Figure 1, we examine the the first decoder layer in CVAE which uses 2 layers Multi-Layer Perceptron(MLP) to build encoder and decoder, and One-Hot vector to represent condition information. We separate the first layer's matrix into $W_c$ with $n1$ columns and $W_p$ with $n2$ columns, in which $n1$ equals to the dimension of Gaussian noise $z$ and $n2$ equals to the dimension of One-Hot vector $c$. As we can see, the result of $W_p \cdot c$ equals the $i^{th}$ column in $W_p$, which $c^i$ is $1$. And the result of $W_c \cdot z$ is the sum of all $W_c$'s columns with different weights. In such decoding process, One-Hot vector excatly performs as a \textbf{selective activator} which activates different paramters in $W_p$. Then the output $(W_c \cdot z + W_p^i + b)$ is passed to a activation layer and then to the second decoder layer, after which final visual data $\hat{x}$ will be generated.

Further more, as $W_c$ is categories independent while $W_p$ is categories dependent, we would like view $W_c$ as a common content generator and $W_p$ a private content generator. The reason that traditional CVAE performs well on generating label-conditioned images even can't minimize the second divergence may that the One-Hot vector is a good replacement of $q_{\phi_y}(y|x^{(i)})$. We may refer to some widely-known neuroscience knowledges that one of the most fascinating functions of our brain is its complex activation and suppression process of different neurons. As the $0$ element in $c$ prevents corresponding parameters from further forward propagation while $1$ permits, One-Hot vector actually a wonderful replacement of $q_{\phi_y}(y|x^{(i)})$.

To briefly summarize, in CVAE structure, latent One-Hot vector input performs as a selective activitor. Decoder performs as a generator that separatedly generates common content and private content. As $z$ is encoded from different categories, encoder performs as a common content recognizer.

\subsection{`Growing' New Condition Neurons}

We have discussed the selective activation function of condition One-Hot input. We continue to use such CVAE's structure as an example to state the dynamically growing process. The encoder receives flattened image vectors and recogizes common content $z \in \mathbb{R}^{n1}$. If not applying continual training process, we need to feed One-Hot vector $c \in \mathbb{R}^{n2}$ (n2 equals total category numbers) to the decoder.

If applying continual training process, as shown in Figure 2, the One-Hot vector's dimension should be equal to the number of already used categories. And the columns of $W_p$ should also equals to the same. As more categories come, we extend the dimension of One-Hot vector as well as adding new columns into $W_p$ (using random initialization). Note that as we extend the columns of private content generator, we are dynamically increasing the neurons in the network that can be actived by the One-Hot vector. And newly growed neurons can then learn to generate the new category's private content.

We further analyse the mathematical properties of such process. Let $z^l$ denote $L^{th}$ layer's output before activation layer, $a^l$ denote $L^{th}$ layer's output after activation layer, and $\delta^l$ denote the gradient of $a^l$. Their relationship with the objective function are shown in equation (4). The relationship between $\delta^{l+1}$ and $\delta^l$ are shown in equation (5). 
\begin{equation}\label{key}
\begin{aligned}
\delta^l = \frac{\partial J(W, b)}{\partial z^{l}} = \frac{\partial J(W, b)}{\partial a^{l}} \cdot \sigma'(z^{l})
\end{aligned}
\end{equation}
\begin{equation}\label{key}
\begin{aligned}
\delta^l = (\frac{\partial z^{l+1}}{\partial z^{l}})^\mathrm{T}\delta^{l+1}  = (W^{l+1})^\mathrm{T}\delta^{l+1} \cdot \sigma'(z^{l})
\end{aligned}
\end{equation}
At each time, only one element isn't zero in the One-Hot vector. So at each backpropagation process, the gradient of $W_p$ has the following property. Let $a_0$ be the One-Hot vector, $W{_p^i}$ be $W_p$'s $i^{th}$ column.
\begin{equation}\label{key}
\begin{aligned}
\frac{\partial J(W, b)}{\partial W_p} = \delta^1(a_0)^\mathrm{T}
\end{aligned}
\end{equation}
Further more,
\begin{equation}\label{key}
\frac{\partial J(W, b)}{\partial W{_p^i}} =
\begin{cases}
\delta^{1i}&\mbox{$a_0^i = 1$}\\
0&\mbox{$a_0^i = 0$}
\end{cases}
\end{equation}

Noting that after backpropagation, we have $W{_p^{i'}} = W_p^i - \alpha \times \frac{\partial J(W, b)}{\partial W{_p^i}}$. Above property tells that because of the selective activation, in backpropagation, each categories-dependent paramters in $W_p$ update independently. However, the updation of $W_c$ is related with all categories. This important property conforms how we understand the term `private' and `common'. And mathematicallty, it supports the correctness of our dynamically growing structure.

\subsection{Losses and Circulatory mode}

Our model contains three losses: traditional CVAE loss, circulatory loss, and self-memory supervision loss.

\textbf{Traditional CVAE Loss.} Applying our CVAE formular shown in equation(3), traditional one-way CVAE minimizes the first KL divergence between encoder recognized $q_{\phi_z}(z|x^{(i)})$ and the assumed latent distribution (i.e., Gaussian Distribution), and maximums the likelihood probability of decoder generated visible data $\hat{x}$. Let $x$ denotes the available data in the newly comming category.
\begin{equation}\label{key}
\begin{aligned}
\zeta_{new} = 
&\frac{1}{2}\sum_{d}^{i=1}(\mu_i^2+\sigma_i^2-log\sigma_i^2-1)\\
&+BCELoss(x_{true}, \hat{x})
\end{aligned}
\end{equation}
Above loss doesn't manage the second KL divergence in our formula. However, as we mentioned above, using the One-Hot vector as a replacement directly is a stable way to represent the latent distribution of condition information because of selective activation. `Stable' means that it won't make the output diverge even is ignored.

\begin{figure}[!h]
	\vskip 0.2in
	\begin{center}
		\centerline{\includegraphics[width=0.667\columnwidth]{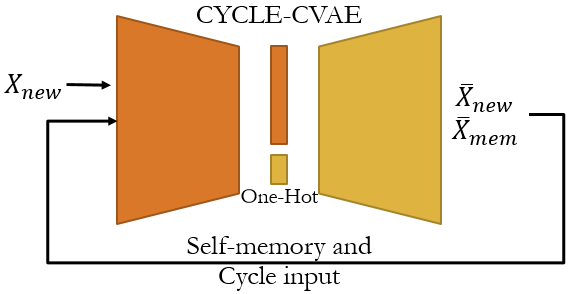}}
		\caption{Self-memory supervision and circulatory process.}
		\label{key}
	\end{center}
	\vskip -0.2in
\end{figure}
\textbf{Self-Memory Supervision and Circulatory Loss.} Figure 3 shows the self-memory supervision and circulatory process. When training newly growing neurons on new categories, we apply self-memory supervision. Before the training start, we call memory of unavailable categories using Gaussian noises and corresponding One-Hot condition vectors. Self-memory supervision has two missions. One is to make the common content recognizer keep the ability to recognize old categories' common content, the other is to make the decoder keep the ability to generate unavailable data.\\
Let $x_{mem}$ be generated unavailable data. $\hat{\mu}_i$,$\hat{\sigma}_i$ are latent representation recognized by encoder with input $\hat{x}_{mem}$.
\begin{equation}
\begin{aligned}
\zeta_{mem} = &\frac{1}{2}\sum_{d}^{i=1}(\hat{\mu}_i^2+\hat{\sigma}_i^2-log\hat{\sigma}_i^2-1)\\
&+BCELoss(x_{mem}, \hat{x}_{mem})
\end{aligned}
\end{equation}
In the traditional forward propagation, we generate visual data $\hat{x}$ and minimize the binary cross-entropy loss(BCELoss). This is a one-way mode because we don't use of those generated images after the calculation of BCELoss. However, because the self-memory supervision process need those generated images as `real' images later, so we need to make sure those images with enough features to accomplish the self-memory supervision. We extent the one-way mode to a circulatory one which means we will re-feed those generated images back to the model as warm-up operation. The circulatory loss is as follows:
\begin{equation}
\begin{aligned}
\zeta_{mem} = &\frac{1}{2}\sum_{d}^{i=1}(\check{\mu}_i^2+\check{\sigma}_i^2-log\check{\sigma}_i^2-1)\\
&+BCELoss(\check{x}, \hat{x})
\end{aligned}
\end{equation}
$\check{\mu}_i$,$\check{\sigma}_i$ are latent representations recognized by encoder with input $\hat{x}$. Because of this circulatory mode, we would like to refer our model as CYCLE-CVAE.

We don't focus on how to improve self-memory supervision mathematically, so three losses have similar formulas. The total loss of our model is:
\begin{equation}
\begin{aligned}
\zeta_{all} = \zeta_{new} + \lambda_1\zeta_{cyc} + \lambda_2\zeta_{mem}
\end{aligned}
\end{equation}
$\lambda_1$ and $\lambda_2$ are set to balance three losses.

\section{Experiments}

In this section, we evaluate our methods on label-conditioned generation using MNIST\footnote{url:http://yann.lecun.com/exdb/mnist/} and Fashion-MNIST\cite{DBLP:journals/corr/abs-1708-07747}. And we compare our results\footnote{More sample results are provided in supplementary material.} with joint learning\footnote{The model is trained on data from all categories.} results using traditional CVAE. Additionally, in above discussions, the separation of common content and private content is a natural production of our methods. So we visualize what the common content looks like and how it changes along the training process. Another natural production is the steady feature of common content, more formally, the regression phenomenon of the common content. It means if the recognizer encodes common content well and the separation between common and private content has been successfully established, memories about old categories won't be lost even not applying self-memory supervision when training on new categories. Both two productions and interesting results are presented bellow.

\textbf{MNIST and Fashion-MNIST:} Before moving on, we would like to state the reason for the choices of datasets. We call MNIST a single pattern dataset inspired by the using of \textit{Seven-Segment Digital Tube}. Ten numbers can all be displayed by activating different segements. So the shape of the tube is what we think the single pattern as well as the common content hiding behind MNIST digits. And for Fashion-MNIST, we can't point out that pattern directly. It contains T-shirt and Trouser(both clothes), Sandal and Ankle boot(both shoes). So we view Fashion-MNIST as a multi-patterns dataset. Single pattern means the existence of meaningful common content while multi-pattern may not. Using these two datasets as comparsion will better evaluate our methods on different perspectives.

\textbf{Training Details:} We choose 2 layers MLP as CYCLE-CVAE's encoder and decoder(Encoder: $784 \rightarrow 256 \rightarrow 2$. Decoder: $(2+n2) \rightarrow 256 \rightarrow 784$, n2 is the dimension of the One-Hot vector). To accomplish dynamically growing, after training on each category, we extend the first decoder layer from $[256, 2+n2]$ to $[256, 2+(n2+1)]$ by copy the trained parameters to corresponding positions and randomly initilize new parameters. We set epochs to $20$ when training models on each category, learning rate to $0.001$ and both $\lambda_1$ and $\lambda_2$ to 1.

\textbf{Quantitative Metrics:} To evaluate dynamically growing process, we focused on comparing this process to joint learning process which applying the same network stucture. We use two metrics: $Acc$ and $r$-$Acc$ following Mengyao Zhai et al.\cite{Zhai_2019_ICCV} to evaluate the generated images' quanlity. $Acc$ is the accuracy using a classifier network trained on real images to evaluate generated images. $r$-$Acc$ is the accuracy using a classifier network trained on generated images to evaluate real images. The closer the two matrics on two processes are, the less side-effect dynamically growing process has.

\subsection{Training Results and Comparision}

First, we conduct experiments on MNIST digits. The visual results of JL and our methods are shown in Figure 4 (the rightmost is JL's results).
\begin{figure}[!h]
	\vskip 0.2in
	\begin{center}
		\centerline{\includegraphics[width=1.1\columnwidth]{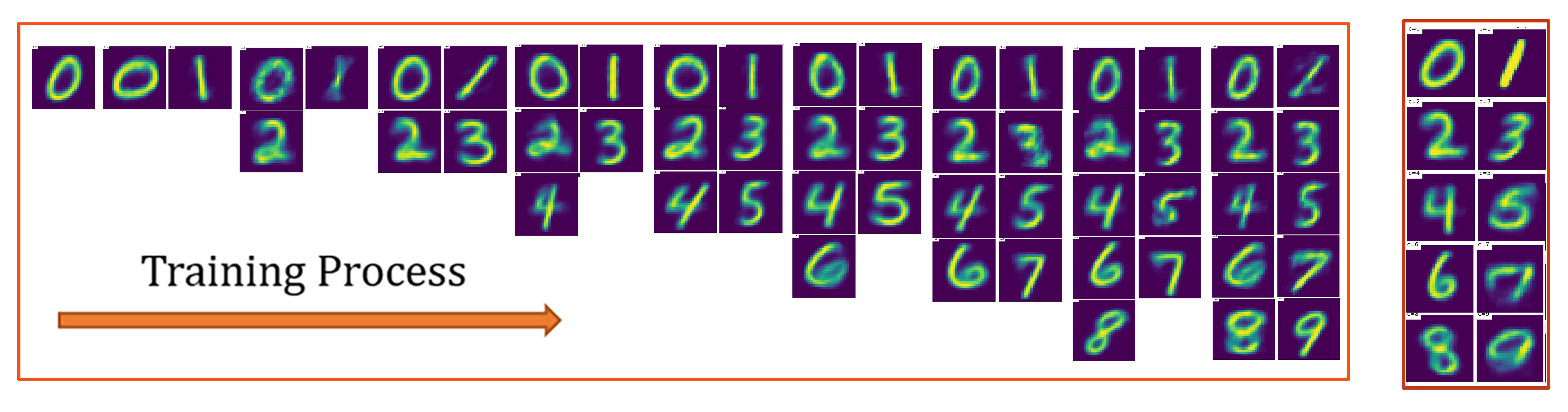}}
		\caption{Training Results and Comparision on MNIST.}
		\label{key}
	\end{center}
	\vskip -0.2in
\end{figure}
\begin{table}[!h]
	\caption{Quantitative evaluation for MNIST digits generation.}
	\label{key}
	\vskip 0.15in
	\begin{center}
		\begin{small}
			\begin{sc}
				\begin{tabular}{cccc}
					\toprule
					 & &JL &Ours\\
					\midrule
					\multirow{2}*{MNIST}& $Acc$ &$99.2\%$& $99.6\%$ \\
					~ & $r$-$Acc$& $83.8\%$ & $83.2\%$\\
					\bottomrule
				\end{tabular}
			\end{sc}
		\end{small}
	\end{center}
	\vskip -0.1in
\end{table}
\\
From the figure, we can see that our methods perform relatively well compared to JL. Meanwhile, VAE is famous about generate samples with different characters (known as data's manifold), we can see our model somehow maintain this ability.\\
The quantitative metrics are shown in Table 1. Two metrics are quite close on two processes which demonstrate growing process doesn't disrupt CVAE's self-encoding ability. Zhai et al.\cite{Zhai_2019_ICCV} also conduct memory playback $(MR)$ and knowledge distillation $(KD)$ on MNIST. The $Acc$ and $r$-$Acc$ of $MR$ are $97.54\%$ and $85.57\%$, and of $KD$ are $97.52\%$ and $87.77\%$. Comparing with $MR$ and $KD$, our methods can generate relatively high quality images from the $Acc$ perspective. However, lacking of discriminator compared with GAN based models makes generated images with less details thus our methods' $r$-$Acc$ is a bit lower.

Second, we conduct experiments on the more challenging Fashion-MNIST dataset. The visual results of JL and our methods are shown in Figure 5 (the rightmost is JL's results).
\begin{figure}[!h]
	\vskip 0.2in
	\begin{center}
		\centerline{\includegraphics[width=1.1\columnwidth]{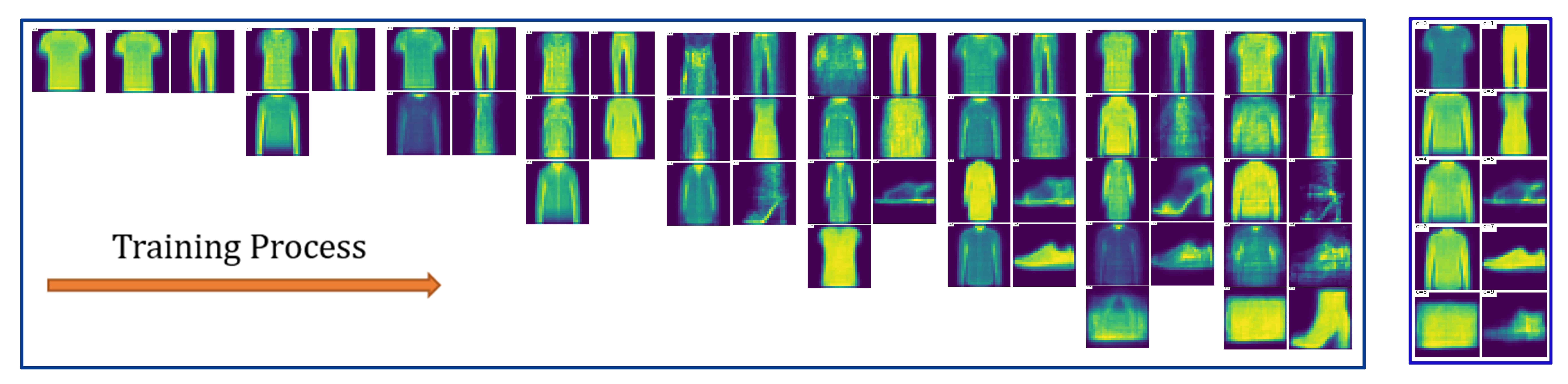}}
		\caption{Training Results and Comparision on Fashion-MNIST.}
		\label{key}
	\end{center}
	\vskip -0.2in
\end{figure}\\
From the results, we can see our methods' results are as quite ideal as we can expect. It distinguishes different categories well but more severe bluriness is observed. Quantitative metrics are shown in Table 2.
\begin{table}[!h]
	\caption{Quantitative evaluation for Fashion-MNIST generation.}
	\label{key}
	\vskip 0.15in
	\begin{center}
		\begin{small}
			\begin{sc}
				\begin{tabular}{cccc}
					\toprule
					& &JL &Ours\\
					\midrule
					\multirow{2}*{Fashion-MNIST}& $Acc$ &$82.1\%$& $81.5\%$ \\
					~ & $r$-$Acc$& $66.1\%$ & $64.2\%$\\
					\bottomrule
				\end{tabular}
			\end{sc}
		\end{small}
	\end{center}
	\vskip -0.1in
\end{table}\\
Comparing to the results on MNIST, two metrics differ larger on Fanshion-MNIST as our expectation that multi-patterns dataset will disturb the effectiveness of learning common content. But they varies in an unsignificant range and dynamically growing process can manage abstract common content as well. This abstract common content we will see in folloing is actually a kind of regression phenomenon.

\subsection{Visualization of Common Content}

In this subsection, we will visualize what common content our models learn when it has been trained on all categories in MNIST. As Fashion-MNIST is a multi-patterns dataset, we are interested about how common content changes along training process, so we keep tracks on it after new categories have been trained.

We would like to first introduce the way to visualize common content. We separate the first decoder layer into $W_c$ and $W_p$. Parameters in $W_p$ are selective activated by One-Hot vector which means each category will active different parameters. If we erase all selected parameters' effects on $W_c \cdot z$, then we may get the pure common content images. In experiments, we replace One-Hot vector by $\vec 0 \in \mathbb{R}^{n2}$ to erase such effects.

\begin{figure}[!h]
	\vskip 0.2in
	\begin{center}
		\centerline{\includegraphics[width=0.14\columnwidth]{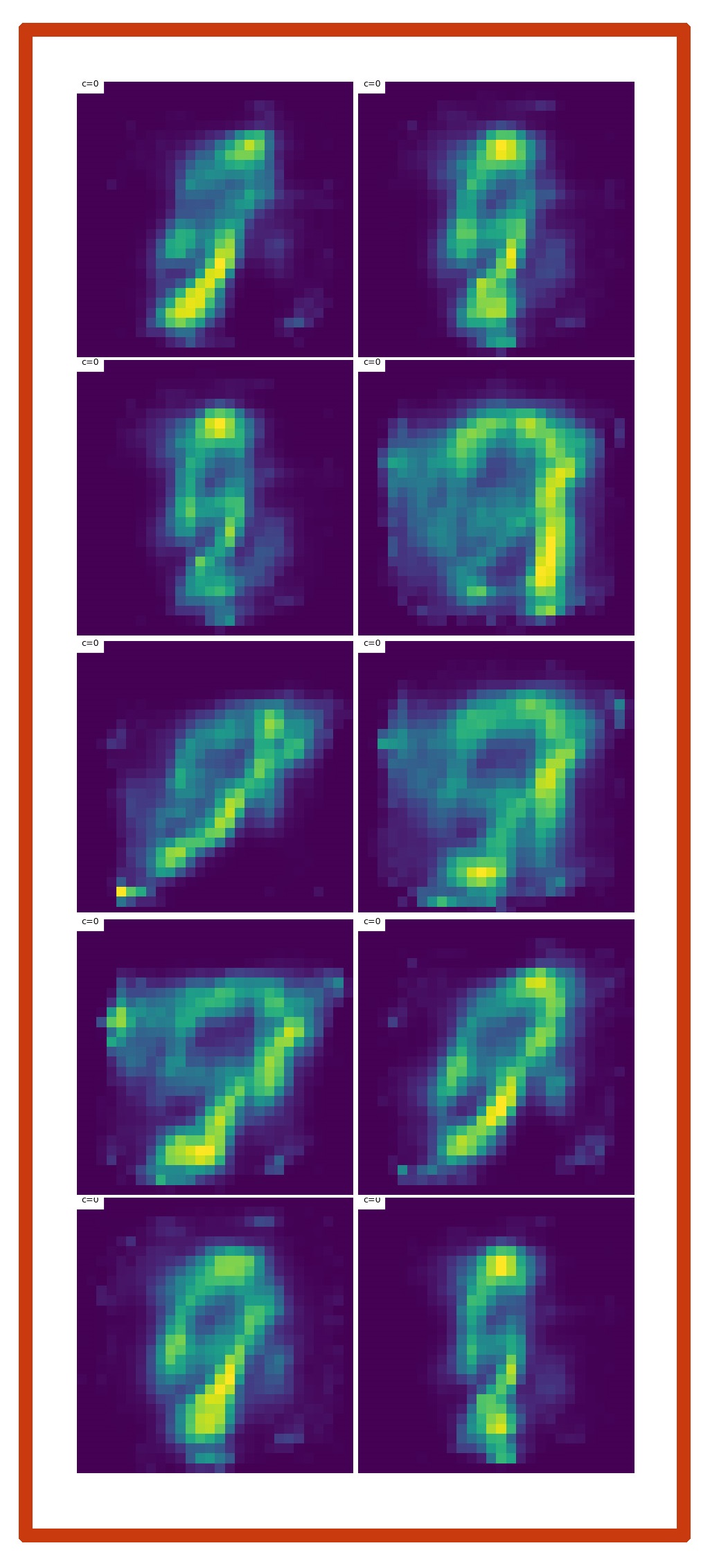}}
		\caption{MNIST Common Content.}
		\label{key}
	\end{center}
	\vskip -0.2in
\end{figure}
In Figure 6, we visulize the common content learnt on MNIST. We may observe each sample with a cycle-like outline and one short bar at the cycle's middle. It corresponds to us Seven Segment Digital Tube inspiration. 

\begin{figure}[!h]
	\vskip 0.2in
	\begin{center}
		\centerline{\includegraphics[width=0.75\columnwidth]{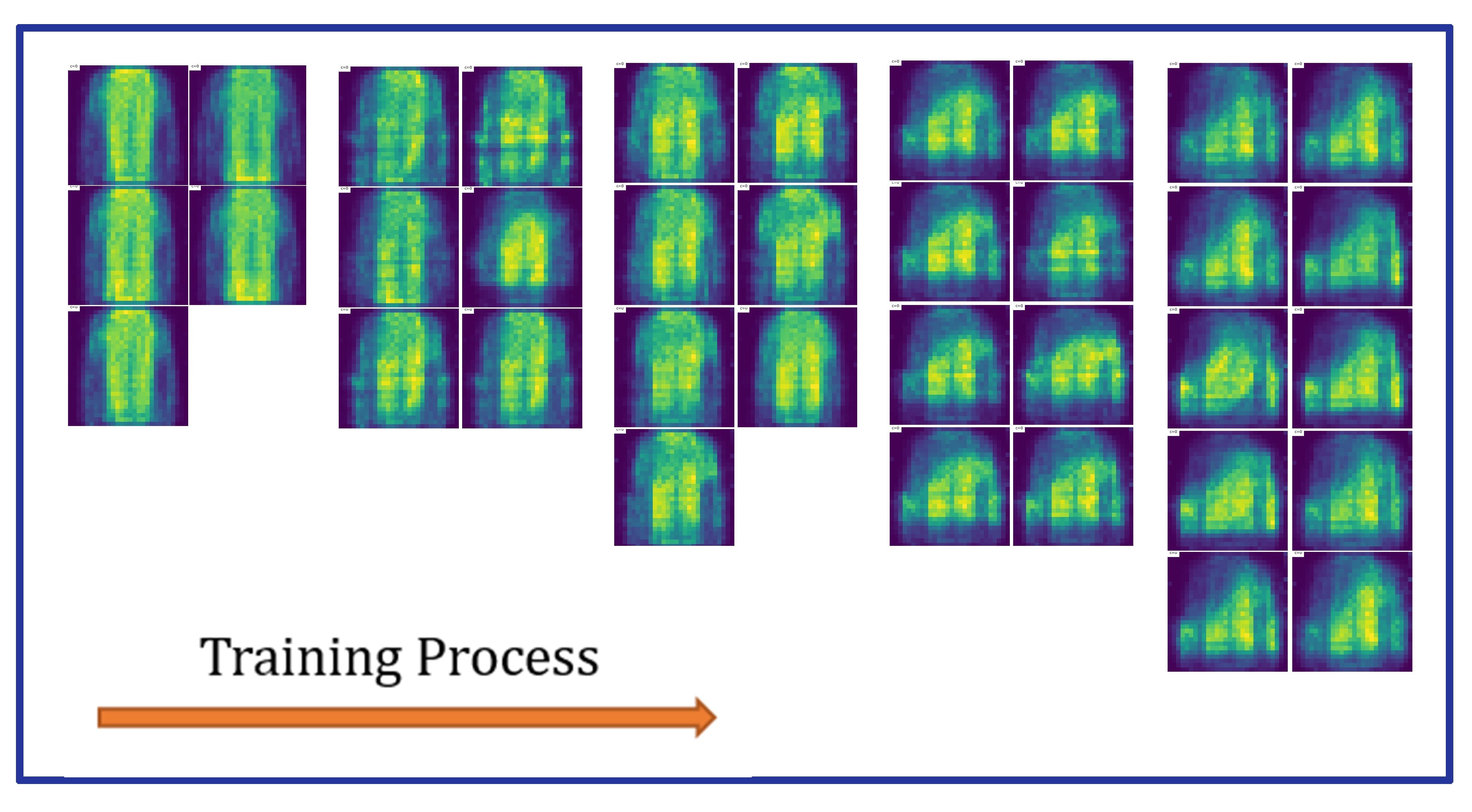}}
		\caption{Fashion-MNIST Common Content.}
		\label{key}
	\end{center}
	\vskip -0.2in
\end{figure}
In Figure 7, each subfigure is one sample of common content\footnote{From left to right, number of already trained categories is 5, 6, 7, 8 and 10.}. When sample the leftmost common content, 5 categories are used and they are T-shirt, Trouser, Pullover, Dress and Coat. We can see the outputs perform high values in the images' middle pixels while low values on each side. Generally, it looks like clothes if we combine each side pixels or looks like Trousers if we ignores those pixels.\\
With the process going on, range of highlight pixels shrink to the middle. Note that Fashion-MNIST contains Sandal, Sneaker and Ankle boot. Those shoes images have high pixel values only in center positions. So the common content seems to highlight those pixels all categories have. Like a regression process, those pixels are the mean of different categories.
\begin{figure}[!h]
	\vskip 0.2in
	\begin{center}
		\centerline{\includegraphics[width=0.25\columnwidth]{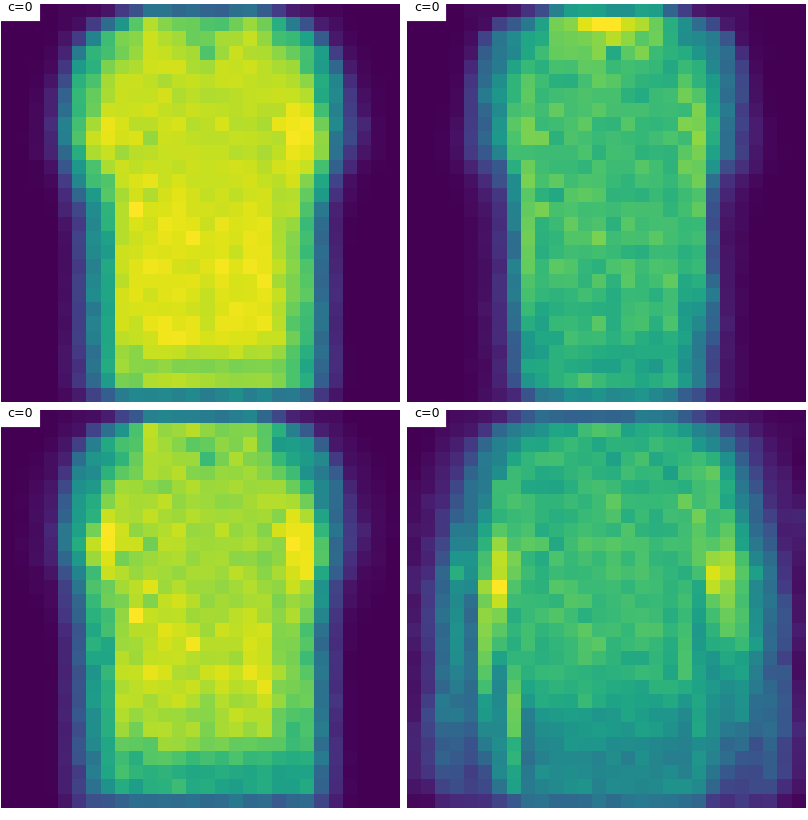}}
		\caption{Common Content on Label 0, 2, 4 and 6.}
		\label{key}
	\end{center}
	\vskip -0.2in
\end{figure}\\
To better convey the idea of common content, as comparision, we conduct experiments on Fashion-MNIST only using images with label 0, 2, 4 and 6 (i.e., T-shirt, Pullover, Coat and Shirt). Common content sample results are shown in Figure 8. Comparing to Figure 7, we can see that as we choose training images with more similar shapes, the common content become more meaningful and recognizable.

\subsection{Steady Feature of Common Content}
\begin{figure}[!h]
	\vskip 0.2in
	\begin{center}
		\centerline{\includegraphics[width=0.14\columnwidth]{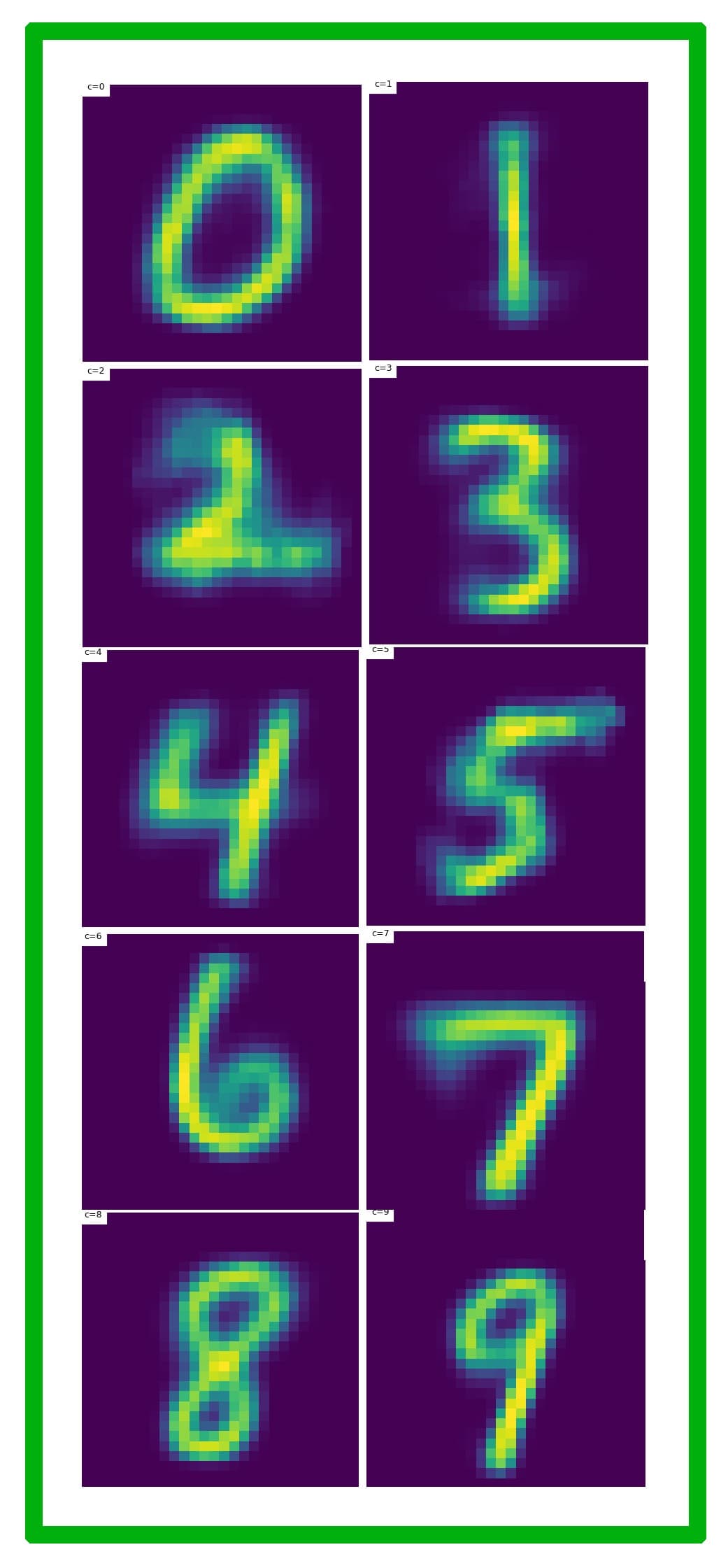}}
		\caption{Training on 9 Without Memory of Number 5.}
		\label{key}
	\end{center}
	\vskip -0.2in
\end{figure}
Finally, we would like to introduce the steady feature of the common content. From our experiments shown in Figure 6, we can conclude our interpretations about common contents and private content are reliable. So may we ask a question: What if we don't conduct self-memory supervision when training on a new category? For example, number 9 contains those common content and each one's private contents are learnt independently. So maybe the answer is we won't confront severe catastrophic forgetting.\\
Figure 9 shows the sample result when we finished training on number 9 without calling memory about number 5. The $Acc$ of such a model is around $99.5\%$, just as same as the result in Table 1. This means with other categories' self-memory supervision, catastrophic forgetting isn't observed on the generation of number 5.
\begin{figure}[!h]
	\vskip 0.2in
	\begin{center}
		\centerline{\includegraphics[width=\columnwidth]{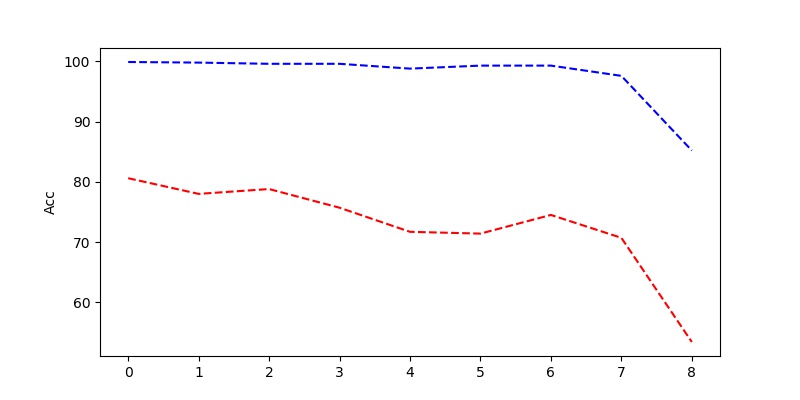}}
		\caption{Acc When lacking Self-Memory Supervision.}
		\label{key}
	\end{center}
	\vskip -0.2in
\end{figure}
\begin{figure}[!h]
	\vskip 0.2in
	\begin{center}
		\centerline{\includegraphics[width=\columnwidth]{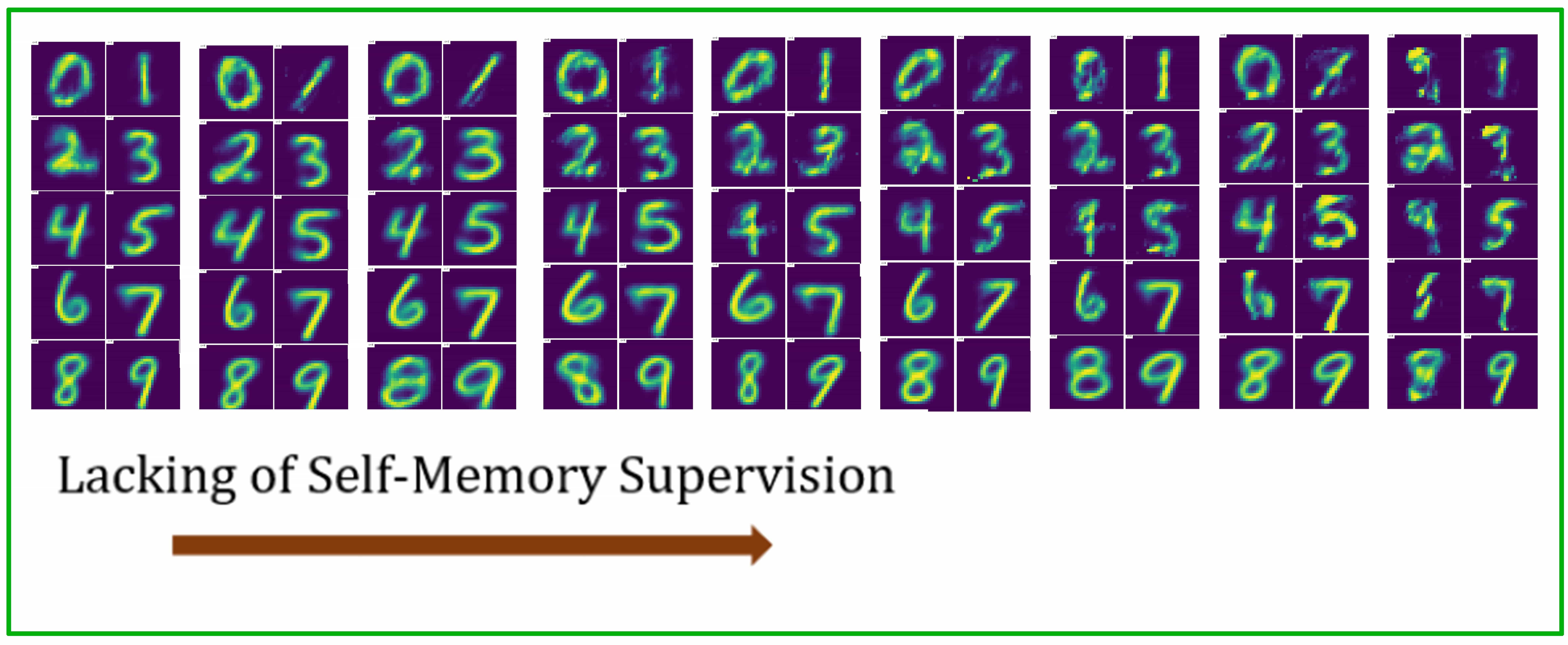}}
		\caption{Memory-Absent Results on MNIST.}
		\label{key}
	\end{center}
	\vskip -0.2in
\end{figure}
\begin{figure}[!h]
	\vskip 0.2in
	\begin{center}
		\centerline{\includegraphics[width=\columnwidth]{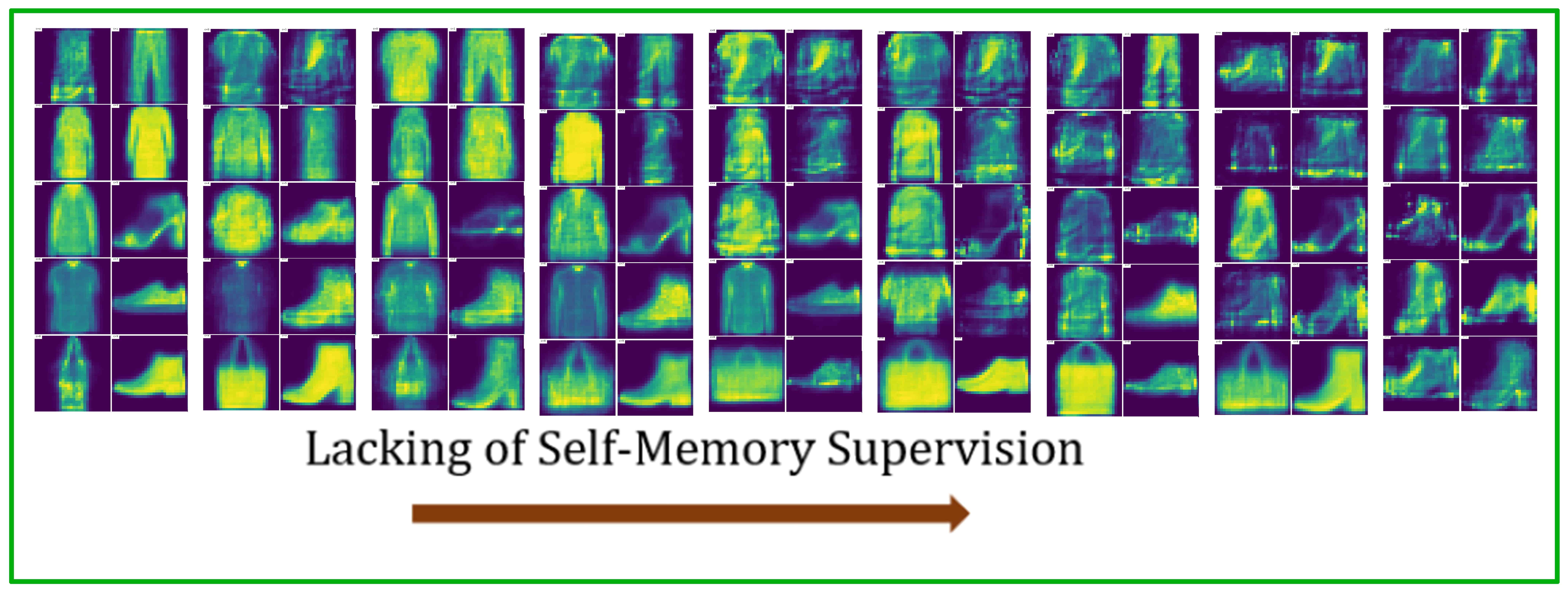}}
		\caption{Memory-Absent Results on FashionMNIST.}
		\label{key}
	\end{center}
	\vskip -0.2in
\end{figure}\\
In Figure 10, we keep tracks on the $Acc$ as we decrease self-memory supervision on more categories when training encoding label 9 (Number 9 in MNIST and Ankle boot in Fashion-MNIST). The x-axis denotes we don't call memory from label 0 to label $x$. Figure 11 and 12 keep tracks on the correspoinding generation results.
As shown in above figures, $Acc$ performs significant decrease when the $x$ comes to $8$ on both dataset, and before that it decreases gradually on Fashion-MNIST while remaining steadily high on MNIST. Along the process, generation results become less recognizable with a low pace. Noting that when $x$ comes to $8$ means no self-memory supervision anymore, we can conclude that other categories are with enough knowledges to force the model keep the ability of encoding and decoding common content even the training data is severe label-absent. While the private content is learnt independently, it brings us above results. We call this the steady feature of common content as it performs high anti-interference when training new categories.

\section{Conclusion}

In this paper, we rewrote the CVAE's formula and focused on the role of condition information. Such rewriting formula allowed us to make a new interpretation about the function of different parts in MLP based CVAE models. The selective activation function of the One-Hot condition vector we pointed out inspired us to propose a dynamically growing network. And combining this network with self-memory supervision process, using simple CVAE models, we were able to accomplish continuingly learning on generation tasks. Besides, we extended traditional one-way CVAE to a circulatory mode to preheat the usage of memory datas. Unlike previous methods that remaing the network structure the same or retraining newly initialized network, our methods made the model dynamically extend itself and could well separate categories' private content from common content. We validated our methods on the single-pattern MNIST and the multi-pattern Fashion-MNIST. Our experiments showed that dynamically growing process won't significantly disturb traditional network's ability. The visualization of common content and the regression phenomenon further verified the correctness of our interpretations mentioned above. The steady feature of common content was also presented.

\bibliography{citation}
\bibliographystyle{icml2020}

\end{document}